\documentclass[letterpaper, 10 pt, conference]{ieeeconf}
\IEEEoverridecommandlockouts   
\usepackage{lipsum} 
\usepackage[T1]{fontenc}
\usepackage{graphicx}
\usepackage{graphics}
\usepackage{multirow}
\usepackage{float}
\usepackage{hyperref}
\usepackage{epstopdf}
\usepackage{enumerate}
\usepackage{multicol,multirow}
\usepackage{booktabs}
\usepackage{cite}
\usepackage{amsmath}
\usepackage{academicons}
\usepackage{amssymb}
\usepackage{amsfonts}
\usepackage{bbm}
\usepackage{balance}
\usepackage{paralist}
\usepackage[binary-units=true]{siunitx}
\graphicspath{{figures/}}
\usepackage[ruled,vlined,linesnumbered]{algorithm2e}
\let\oldnl\nl
\newcommand{\nonl}{\renewcommand{\nl}{\let\nl\oldnl}}

\graphicspath{{figures/}} 

	\title{\LARGE\bf
	   Theoretical Evidence Supporting Harmonic Reaching Trajectories 
	}
	\author{Carlo Tiseo, Sydney Rebecca Charitos, and Michael Mistry

		\thanks{Carlo Tiseo, Sydney Rebecca Charitos and Michael Mistry are with the Edinburgh Centre for Robotics, Institute of Perception Action and Behaviour, School of Informatics, University of Edinburgh. Email: \texttt{carlo.tiseo@ed.ac.uk}}
	   \thanks{This work has been supported by the following grants: EPSRC UK RAI Hubs ORCA (EP/R026173/1), NCNR (EPR02572X/1) ; and EU Horizon 2020 project  THING (ICT-2017-1).}
	}
\begin{document}
\thispagestyle{empty}
\fbox{
\parbox{\textwidth}{
© 2021 IEEE.  Personal use of this material is permitted.  Permission from IEEE must be obtained for all other uses, in any current or future media, including reprinting/republishing this material for advertising or promotional purposes, creating new collective works, for resale or redistribution to servers or lists, or reuse of any copyrighted component of this work in other works.}}
\newpage
\maketitle
\thispagestyle{empty}
\pagestyle{empty}

\begin{abstract}
Minimum Jerk trajectories have been long thought to be the reference trajectories for human movements due to their impressive similarity with human movements. Nevertheless, minimum jerk trajectories are not the only choice for $C^\infty$ (i.e., smooth) functions. For example, harmonic trajectories are smooth functions that can be superimposed to describe the evolution of physical systems. This paper analyses the possibility that motor control plans using harmonic trajectories, will be experimentally observed to have a minimum jerk likeness due to control signals being transported through the Central Nervous System (CNS) and muscle-skeletal system. We tested our theory on a 3-link arm simulation using a recently developed planner that we reformulated into a motor control architecture, inspired by the passive motion paradigm. The arm performed 100 movements, reaching for each target defined by the clock experiment. We analysed the shape of the trajectory planned in the CNS and executed in the physical simulator. We observed that even under ideal conditions (i.e., absence of delays and noise) the executed trajectories are similar to a minimum jerk trajectory; thus, supporting the thesis that the human brain might plan harmonic trajectories. 
\end{abstract}

\IEEEpeerreviewmaketitle

\section{Introduction}
Human reaching trajectories have been widely studied for understanding human motor control, laying the foundation not only of rehabilitation robotics but also influencing trajectory planning in robotics. Thus, improving our understanding about these mechanisms can contribute to designing better robot and rehabilitation therapies for people suffering from motor impairments.

\textit{ Flash and Hogan} identified minimum jerk trajectories as the closest to the hand tangential trajectories in human movements \cite{flash1985}. This evidence has determined the deployment of minimum jerk trajectories to compute the ''Cost to Go'' that is used by the basal ganglia as a cost function for identifying the optimal strategy in the optimal motor-control theory \cite{shadmehr2008computational,diedrichsen2010coordination,shadmehr2005computational}.  However, this motor-control theory is incompatible with the information delays associated with neural transmission. The Passive Motion Paradigm (PMP) has been proposed to overcome this issue by using a passive impedance controller which is robust to information delays \cite{ivaldi1988kinematic,mohan2011passive,tommasino2017, tommasino2017task, tiseo2018thesis}. A similar approach has been recently used to define the inclusion of dynamic primitives via the optimisation of multiple impedance controllers connected in parallel\cite{Averta2020,Nah2020}. Nevertheless, all these models still rely on the minimum jerk for planning task-space trajectories.   
\begin{figure}[!htbp]
\centering
\includegraphics[width=\columnwidth]{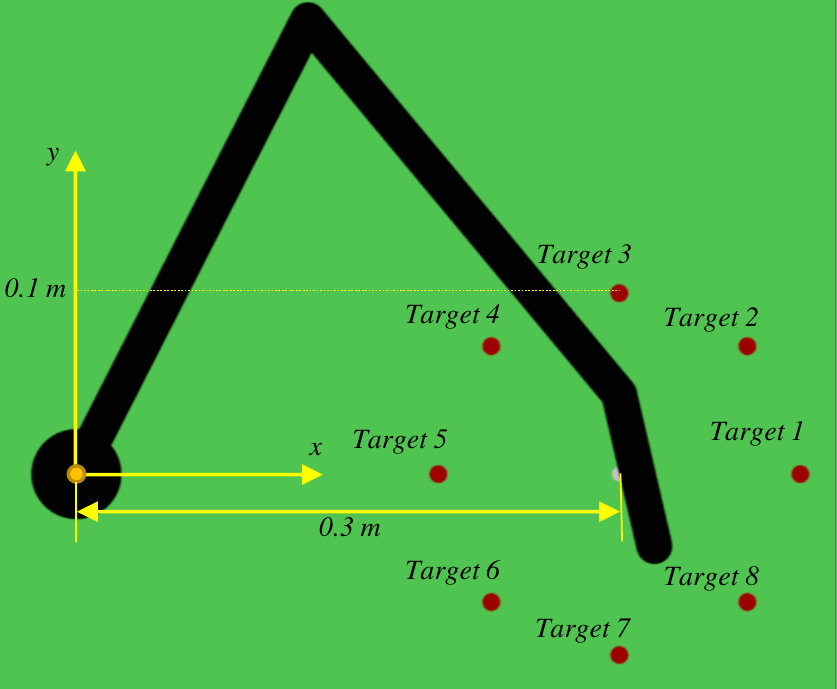}
\caption{Top view of the simulator used to validate that reaching movements are distorted harmonic trajectories. It shall be noted that the hand end-effector is placed on the inner side of the third link.}
\label{fig:SimSetup}
\end{figure} 

\begin{figure*}[!htbp]
\centering
\includegraphics[width=16cm, trim=0cm 1.5cm 0cm 2cm, clip]{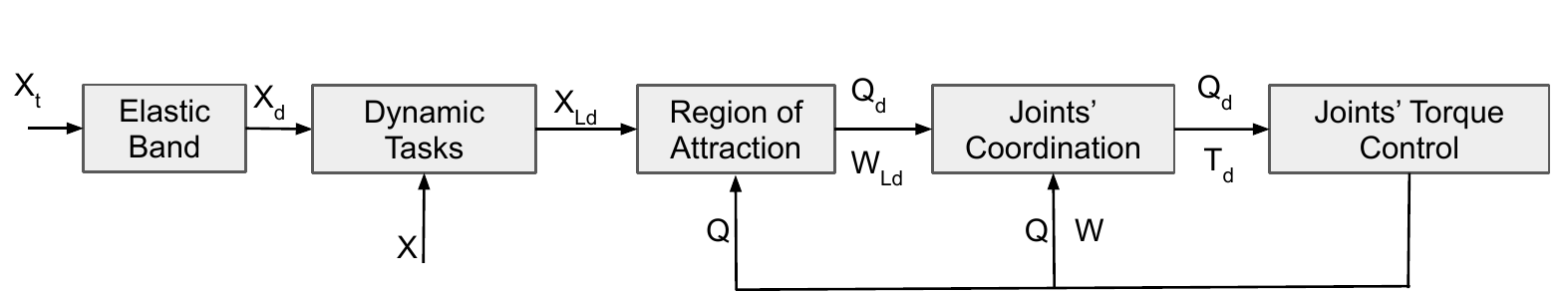}
\caption{The H-PMC is hierarchical architecture taking a desired target state in input and transforming into an action via a cascade of controllers. The elastic band pulls the end-effector along the harmonic trajectory. The dynamic task adds an internal model of the task using the end-effector position error ($X_\text{d}-X$) and the unit vector of the expected interaction direction ($W_\text{t}$). Its outputs are the end-effector poses ($X_\text{Ld}$) for every link of the kinematic chain. The region of attraction generates the postural control $Q_\text{d}$ by identifying the joints' configuration and the desired wrenches for every link in the chain ($W_\text{Ld}$), based on the current pose ($Q$). The joints' coordination uses an haptic controller that transform the $W_\text{Ld}$ in joints' torques ($T_\text{d}$) which are passed together with $Q_\text{d}$ to the joints' controllers.}
\label{fig:CtlArc}
\end{figure*} 

Minimum jerk trajectories are smooth (i.e., $\mathcal{C}^\infty$) polynomial functions. Nevertheless, there are other functions (polynomial and non) that are smooth. \textit{M. J. Richardson et al in \cite{richardson2002comparing}} also studied other alternative polynomial trajectories to replicate human movement. Harmonic trajectories are smooth functions \cite{ji2017riemann}, which describe the autonomous trajectories used by physical systems for transferring energy. The natural frequency of a system refers to the frequency that is preferred for information transferal. Thus, it determines oscillation of an isolated system after perturbation. For example, acoustic dissipation has a natural frequency equal to the musical note associated with it. The natural frequency (i.e., cut-off frequency) also indicates the maximum frequency that information can propagate through a system without information loss. In fact, the system transfer functions show a reduction in amplitude and a shift in phase beyond the system's natural frequency \cite{MillmanMicroelectronics, isidori2013nonlinear}. The Fourier's series also guarantees that any smooth function can be described as a superimposition of harmonic functions \cite{oppenheim2001discrete}, making them an ideal candidate for representing the primitive shape of biological movements.  The Central Pattern Generator (CPG) provides further support to the possibility that harmonic functions are the primitive trajectories used for the generation of human movements. CPGs are specialised neural network oscillators found in the spinal cord, which are responsible for the generation of rhythmic motions and movement coordination \cite{Hogan2013,Bucher2015}. The output of a CPG is a modulated harmonic function, making them variable frequency oscillators.

This manuscript proposes a Harmonic Passive Motor Control (H-PMC) architecture that uses harmonic trajectories as motor primitives based on the PMP. A similar architecture was theoretically proposed in \cite{tiseo2018thesis}, but it was not feasible due to the lack of an adequate controller. The recently developed Fractal Impedance Controller \cite{babarahmati2019} enables the development of the hierarchical architecture of semi-autonomous controllers. The architecture was initially proposed, to the best of the authors' knowledge, for locomotion in \cite{ahn2012walking}. 

\section{Method}
\subsection{Fractal Impedance Controller}
The controller uses a fractal attractor to ensure passive asymptotically stable behaviour. Multiple FIC's can be combined both in parallel and in series without affecting the system stability \cite{tiseo2020}, meaning the controller is able to plan and deal with redundant degrees of freedom. This marks a change from the method proposed by \cite{Averta2020}, since this system does not need to solve a numerical optimisation problem to coordinate multiple controllers without affecting their stability \cite{tiseo2020Planner, tiseo2020}. We have also recently proved that the system can act as a soft admittance controller to interact with unknown external dynamics  \cite{tiseo2020bio}. Another benefit of the FIC is its conservative energy, making it intrinsically robust to significant information delay, and reducing communication bandwidth between the architecture components \cite{babarahmati2019,babarahmati2020}. 

The controlling command from the FIC is determined using the following equation \cite{tiseo2020bio}:
\begin{equation}
\tau=\left\{\begin{array}{cc}
     J^T F_\text{FIC} \left( \tilde{x}\right) &  \text{divergence}\\
     J^T \frac{2F_\text{FIC}\left(\tilde{x}_\text{max}\right)}{\tilde{x}_\text{max}} \left( \tilde{x}-\frac{\tilde{x}_\text{max}}{2} \right)&  \text{convergence}
\end{array} \right.
    \label{FICBasic}
\end{equation}
where $F_{FIC}$ is a generic continuous finite effort (force/torques) profile, $\tilde{x}$ is the end-effector distance from the desired pose and $\tilde{x}_\text{max}$ is the maximum distance reached during the divergent phase. 
\subsection{Harmonic Passive Motor Control}
The architecture in \autoref{fig:CtlArc} shows an input target $x_\text{t}$ in the task-space and generates a smooth optimised movement to reach it. 

\subsubsection{Elastic Band}
The elastic band is implemented, with a more detailed description of its function is available in \cite{tiseo2020Planner}. It uses a Model Predictive Control (MPC) architecture to project the information forward using the desired task-dynamics. A 2D linear elastic band on the \textit{xy}-plane is implemented in this manuscript, as follows:
\begin{equation}
X_d=\iint_{t_\text{i}}^{t} \ddot{X}_\text{d} dt^2=\iint_{t_\text{i}}^{t} \frac{K_\text{d}\left(X_\text{d}\left(t-1\right)-X_\text{t}\right)}{M_\text{d}} dt^2
    \label{EqElasticBand}
\end{equation}
where $X_\text{t}$ is the current target location, $K_\text{d}$ is the stiffness of the elastic band  and the $M_\text{d}$ is the desired inertia for the end-effector. $K_\text{d}$ is determined FIC algorithm by rewriting  \autoref{FICBasic} as a non-linear stiffness centred in the desired pose as shown in \cite{babarahmati2019}. We opted for a linear force profile with upper-bounded force magnitude ($F_\text{max}$) for this controller. The value of $F_\text{max}$ is determined by the product of $M_\text{d}$ and the maximum desired acceleration ($a_\text{max}$).

\subsubsection{Dynamic Tasks}
The dynamics tasks use the current end-effector pose $X$ and the expected direction of interaction for the task $W_\text{t}$ to maximise the orthogonality between the tangent-space of the kinematic chain with the expected direction of the perturbation $W_\text{t}$. The output of this block is the end-effector pose for each link ($X_\text{Ld}$) or the equivalent joint configuration. This can be completed by solving the optimisation problem:
\begin{equation}
    \label{optmin}
    \begin{array}{l}
         \min\left(W_\text{t}^\text{T}\left(t\right)J\left(Q\left(t\right)\right)J^T\left(Q\left(t\right)\right)W_\text{t}\left(t\right)\right)  \\
         s.t.\\
         ~~~X\left(t\right)=X_\text{d}\left(t\right)\\
         ~~~|Q\left(t\right)- Q\left(t-dt\right)| \le \Delta Q_\text{m}
    \end{array}
\end{equation}
where $J$ is the geometric Jacobian of the kinematic chain and $\Delta Q_\text{m}$ is the maximum arc that each joint can move during $dt$. However, this optimisation problem has an algebraic solution for a 3-link arm that entails solving the inverse kinematics using the orientation of the force as the orientation of the desired end-effector. The wrist end-effector target pose is then defined as:
\begin{equation}
    \label{optminSolWrist}
    \begin{array}{l}
         X_\text{W}=X_\text{d}-l_\text{H}
              \left[\begin{array}{c}
              \cos\left(\phi _\text{Wt}\right) \\
              \sin\left(\phi _\text{Wt}\right)
              \end{array}\right]\\
    \end{array}
\end{equation}
where $l_\text{H}$ is the distance from the hand of the end-effector to the wrist joint and $\phi_\text{Wt}=\text{atan2}\left(W_\text{t}(2),W_\text{t}\left(1\right)\right)$. 

\begin{equation}
    \label{optminArm}
    \begin{array}{l}
         X_\text{A}=l_\text{A}
              \left[\begin{array}{c}
              \cos\left(\phi _\text{A}\right) \\
              \sin\left(\phi _\text{A}\right)
              \end{array}\right]\\
        \phi _\text{A}= \phi_\text{Wt} \pm \left|\tan^{-1}\left(\alpha\left(2\right)/\alpha\left(1\right)\right)\right|\\
        \alpha= \left[\begin{array}{c}
              \left(2l_\text{A}\left|\left|X_\text{W}\right|\right|^2\right)^2 -\left(\left|\left|X_\text{W}\right|\right|^2-l_\text{A}-l_\text{FA}^2\right)^2\\
              \left(\left|\left|X_\text{W}\right|\right|^2-l_\text{A}-l_\text{FA}^2\right)^2
              \end{array}\right]
    \end{array}
\end{equation}
where $l_\text{A}$ and $l_\text{FA}$ are the lengths of the arm and the forearm. 
\subsubsection{Region of Attraction}
\label{RAsub}
The system generates a region of attraction for each end-effector of the 3 links (arm, forearm and hand) by superimposing the non-linear FIC controllers as in \cite{tiseo2020}. The non-linear force profile used for the impedance is:
\begin{equation}
    \label{FICRA}
    \begin{array}{l}
         F=\left\{\begin{array}{ll}K_0\tilde{x}=
              K_0 \left(x_\text{d}-\mathcal{K}\left(Q\right)\right),& \tilde{x}\le 0.95\tilde{x}_\text{b} \\
              \frac{\Delta F\left(\tanh\left(\frac{\tilde{x}-\tilde{x}_\text{b}}{0.1353\tilde{x}_b}+\pi\right)+1\right)+F_0}{2},& \text{o/w}
              \end{array}\right.\\
    \end{array}
\end{equation}
where $K_0\le F_\text{max}/\tilde{x}_\text{b}$ is the constant stiffness, $\mathcal{K}(Q)$ is the direct kinematics, $\Delta{F}=F_\text{max}-F_0$, $F_0=0.95K_0\tilde{x}_\text{b}$, and $\tilde{x}_\text{b}$ is the position error at force saturation.

\subsubsection{Joints' Coordination}
This component is the haptic module that tracks the desired end-effector wrench as follows:
\begin{equation}
\label{JCoordination}
    T_\text{d}=J(Q)^\text{T}(2W_\text{d}-W);
\end{equation}
where $T_\text{d}$ are the desired torques, $W_d$ the desired wrench and W the wrench measured at the end-effector. The coordination with the other task-space controller is achieved through the online tuning of the maximum torques of the joint space controllers ($T_\text{max}$), which are derived as follows:
\begin{equation}
    \label{TauMax}
    T_\text{max}=\left[\begin{array}{c}
         \min\left(T_{Amax}\left(1\right),\left|J_\text{1}^\text{T}\left(Q\right)W_\text{Ad}\right|\right)\\
         \min\left(T_{Amax}\left(2\right),\left|J_\text{2}^\text{T}\left(Q\right)W_\text{FAd}\right|\right)\\
         \min\left(T_{Amax}\left(3\right),2\left|J_3^\text{T}\left(Q\right)W_\text{d}\right|\right)
    \end{array}\right]
\end{equation}
where $T_{Amax}$ are the maximum actuation torques, $J_\text{i}^{T},~i=1,2,3$ are the first, second and third row of the geometric Jacobian transposed. $W_\text{Ad}$ and $W_\text{FAd}$ are the desired end-effector wrenches for the arm and forearm, respectively, that have been received as input from the region of attraction. 
\subsubsection{Joints' Torque Controllers}
The implementation of the joints' torque controllers is similar to its use in the region of attraction presented in section \ref{RAsub}. However, their torques profiles have two plateaus, and the tracking error is computed directly by the difference from the desired and the measured joint angle. The first plateau occurs at the desired torque, and the second for the saturation at the maximum joint torque. The additional plateau has been added to guarantee the desired torque, until the tracking error is considered acceptable for the task. Currently, these values are static but they can be adjusted online to alter the arm's mechanical properties for different task requirements. 

\subsection{Experiment Design}
\subsubsection{Simulator}
A simulation of the clock experiment using a planar 3-Link arm has been implemented in Simulink (Mathworks, USA) using the Simscape library. The solver used for the simulation is a \texttt{ode45} with time-steps in the range $[10^{-5}, 10^{-3}]~\si{\second}$. All the output signals are sampled at \SI{1}{\kilo\hertz}. The centre of the clock is located at $[0.3,0]~\si{\meter}$ and the eight targets were equally spaced around a circumference with a radius of \SI{0.1}{\meter}. The length of the links are $l_\text{A}=\SI{0.282}{\meter}$, $l_\text{FA}=\SI{0.269}{\meter}$ and $l_\text{H}=\SI{0.044}{\meter}$. The masses of the links are $M_\text{A}=\SI{4}{\kilo\gram}$, $M_\text{FA}=\SI{2.5}{\kilo\gram}$ and $M_\text{H}=\SI{1}{\kilo\gram}$. We have run 200 reaching movements for each target, which include 100 Home to Target and 100 Target to Home movements. The targets were updated at a frequency of \SI{1}{\hertz}. 

\subsubsection{Data Analysis}

The ratio between the average and the peak velocity ($r$) has been used to compare the planned trajectories $X_\text{d}$ and the executed trajectories $X$. The coefficient was initially used in \cite{flash1985}, and more precisely explained in \cite{richardson2002comparing}, representing the ratio between the peak and the average power. The $r$ values from the simulation trajectories are compared with the human data reported in \cite{flash1985}, a theoretical minimum jerk trajectory r$_\text{mj}=1.875$ and a harmonic trajectory (r$_\text{h}=1.596$). We have also analysed the tracking accuracy of the trajectories computing the Root Mean Square Error (RMSE) between $X_\text{d}$ and $X$ for both position and velocity data. We will compare the tracking of the trajectory and the velocity profiles for the different targets, and compare it with the arm manipulability ellipsoid \cite{siciliano2010robotics}. Finally, we will compare the trajectory data for position and velocity against minimum jerk ($x_{\text{mj}}$) and harmonic ($x_{\text{h}}$) trajectories reported below.
\begin{equation}
    \label{HMJFormulation}
\begin{array}{c}
         x_{mj}\left(t\right)=D\left(10\left(\frac{t}{\Delta t}\right)^3-15\left(\frac{t}{\Delta t}\right)^4+6\left(\frac{t}{\Delta t}\right)^5\right)\\\\
         x_{h}(t)=\frac{D}{2}\left(sin\left(\frac{\pi}{\Delta t}t-\frac{\pi}{2}\right)+1\right) 
\end{array}
\end{equation}
where $\Delta t$ is the movement duration.
\section{Results}
\begin{figure}[!htb]
\centering
\includegraphics[width=\columnwidth, trim=6cm 11.1cm 6cm 10.5cm, clip]{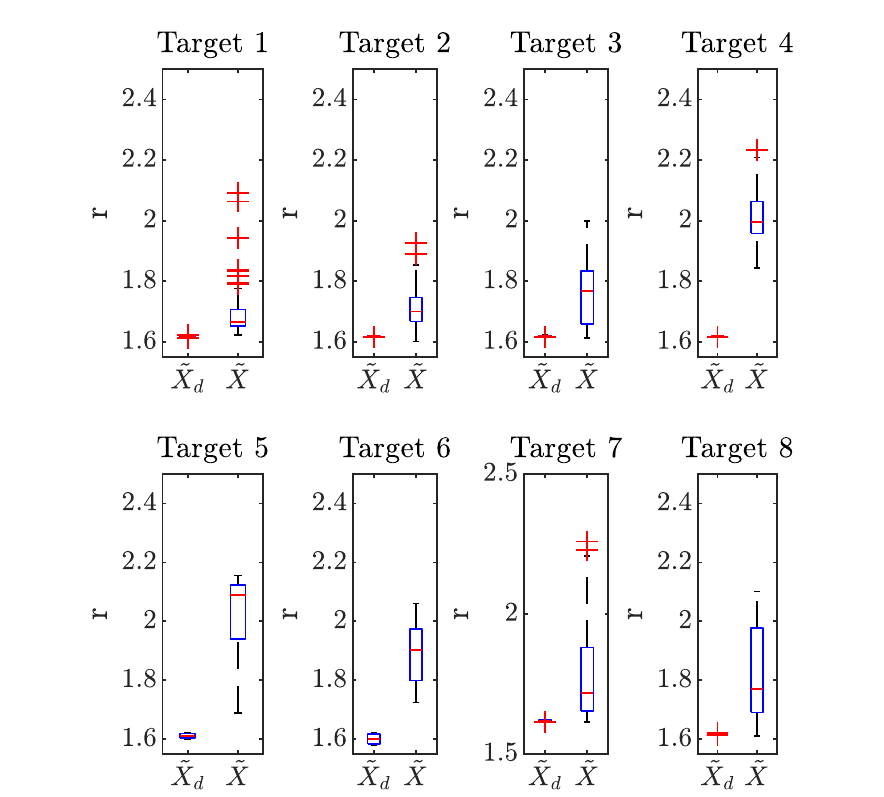}
\caption{The r values for planned trajectories ($X_\text{d}$) have values close to an harmonic trajectory (r$_\text{h}=1.596$), compared against the end-effector trajectories showing values closer to human values (r$_\text{hd}=1.805$) and the minimum jerk (r$_\text{mj}=1.875$) \cite{flash1985}.}
\label{fig:8targetsr}
\end{figure}

\begin{figure}[!htbb]
\centering
\includegraphics[width=\columnwidth,trim=6cm 12cm 6cm 11.75cm, clip ]{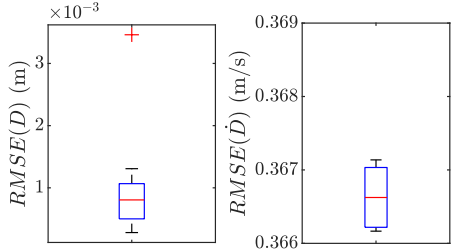}
\caption{The RMSE values show that a high difference in the velocity tracking does not have an evident effect on the trajectory tracking accuracy.}
\label{fig:8targetsRMSE}
\end{figure}

The r-value recorded from the $1600$ trajectories is $1.616\pm0.007$ for $x_d$ and is $1.84 \pm 0.127$ for the end-effector trajectories, which is close to the human movement value of r$_\text{hd}=1.805\pm0.153$ reported in \cite{flash1985}. \autoref{fig:8targetsr} shows the values for the single targets. The figure indicates that the r values are related to the direction of motion due to the anisotropic characteristics of the mechanical structure, which have a preferable direction of motions—meaning Targets 5 and 6 are the more difficult to reach. The RMSE for the trajectories and the RMSE for the velocity are reported in \autoref{fig:8targetsRMSE}. The figure shows that altering the trajectories velocity does not have an impact on the systems trajectory tracking accuracy.
\begin{figure}[!htbp]
\centering
\includegraphics[width=\columnwidth,trim=6cm 9.8cm 6cm 10.1cm, clip ]{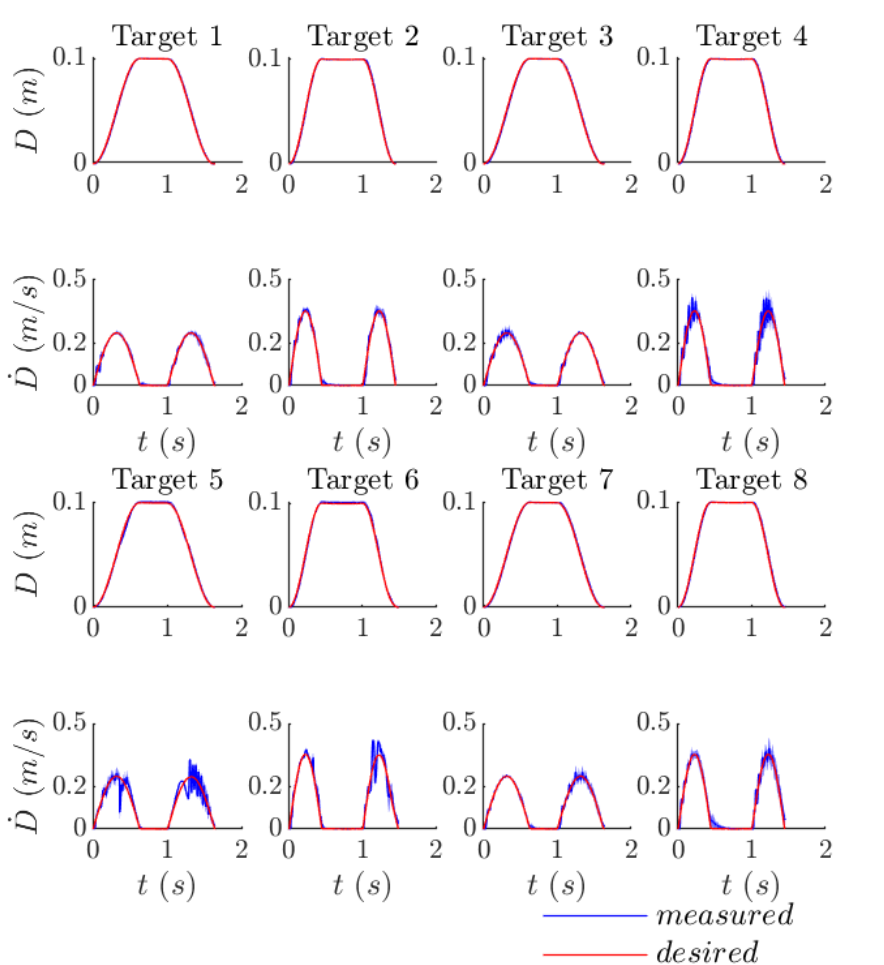}
\caption{The end-effector's trajectories and velocity profiles along the segments which connect the "home" position with the 8 targets that are shown. They confirm that inaccuracies in velocity tracking do not affect the end-effector position accuracy. The velocity errors also indicate that they are directional, increasing in the configurations with lower manipulability. We also observe behaviours consistent with the submovement theory proposed in \cite{morasso1982trajectory}, which is evident in the velocity profiles of Targets 5 and 6. }
\label{fig:TrajectoryComparison}
\end{figure}
\begin{figure}[!htb]
\centering
\includegraphics[width=\columnwidth,trim=6cm 10.9cm 6cm 11.4cm, clip ]{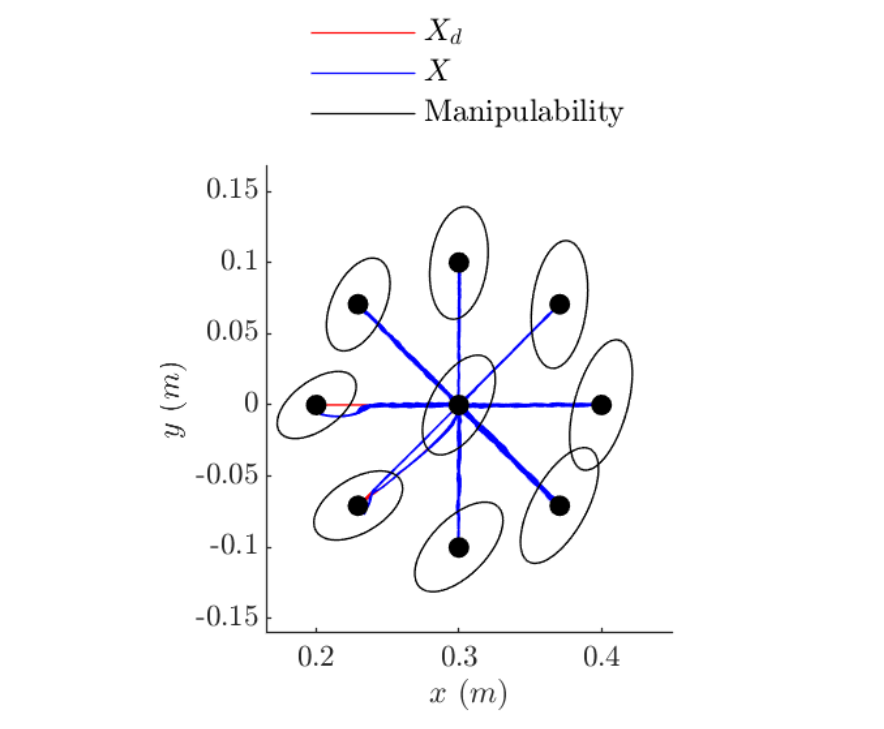}
\caption{The analysis of the manipulability in the home and the target positions confirms the relationship with tracking performances. Showing both orientation and the length of the ellipsis axis affects the performances.}
\label{fig:ClockTrj}
\end{figure}
\begin{figure}[!htbp]
\centering
\includegraphics[width=\columnwidth,trim=6cm 10.75cm 5.5cm 10.25cm, clip ]{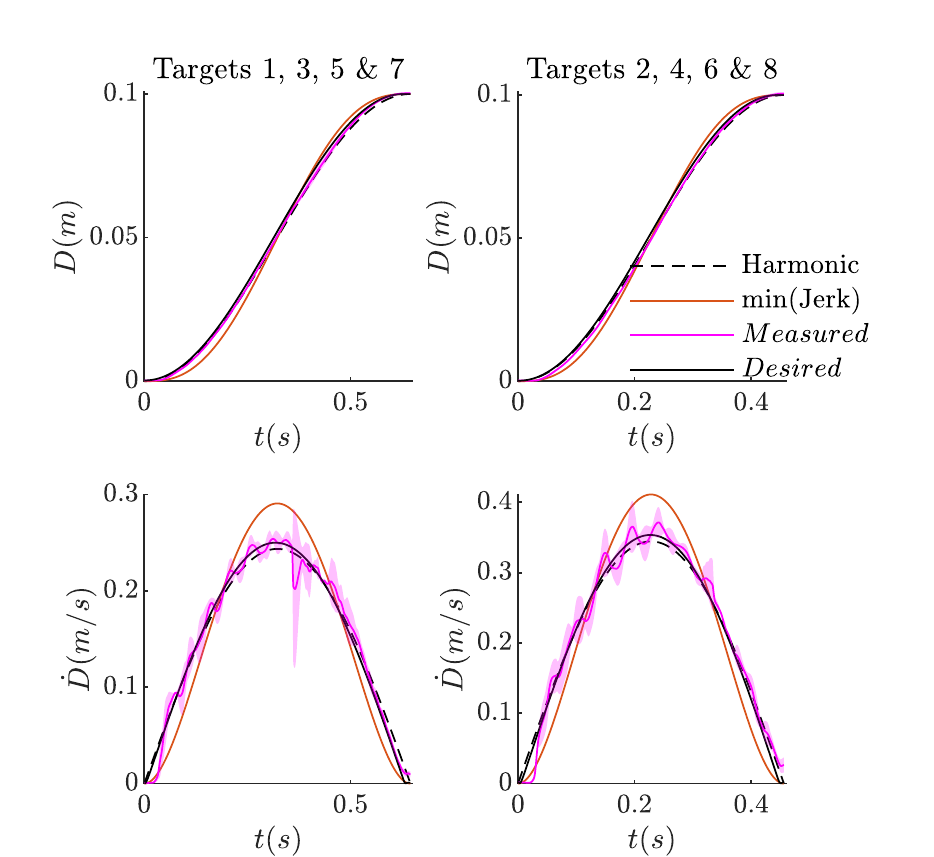}
\caption{The movements data compared with harmonics and minimum jerk trajectories. Despite the position data are almost indistinguishable, the difference is noticeable in the velocity profiles showing that our movements are distorted harmonics rather than minimum jerk.}
\label{fig:MJHComp}
\end{figure} 
The analysis of the trajectories and velocity profiles, described in \autoref{fig:TrajectoryComparison}, confirms that the tracking error in velocity observed in \autoref{fig:8targetsRMSE}. It also indicates that the velocity tracking error is directional and related to the anisotropic behaviour of the kinematic chain. In fact, the tracking worsens when moving to an area with lower manipulability, as reported in \autoref{fig:ClockTrj}. 

\autoref{fig:MJHComp} shows the comparison between the recorded data and the harmonic and minimum jerk trajectories as reported in \autoref{HMJFormulation}. The position data indicates that the trajectories are very similar. On the other hand, the velocity data clearly shows that the measured trajectories are closer in shape to a harmonic trajectory and despite the distortion, their peaks are smaller than the equivalent minimum jerk trajectory.

\section{Discussion}
The results show that the proposed method is capable of generating smooth reaching trajectories with $r$-values in the range of human reaching movements reported in literature \cite{flash1985}. The tracking accuracy is lower than \SI{1}{\milli\metre}, but the velocities are distorted due to the kinematic chain. The distortion is clearly connected to the manoeuvrability of the system, and it is predominant in the regions with lower manipulability. Nevertheless, it remains clear that the executed trajectories are closer to a harmonic rather than a minimum jerk trajectory. The velocity data also show an emergent behaviour that seems consistent with submovements when going through difficult manoeuvres (e.g., Target 5 and 6 in \autoref{fig:MJHComp}), but a more systematic analysis is required to confirm it. 

The proposed control structure is composed by a hierarchical architecture of semi-autonomous controllers, made possible by the properties of the FIC. The elastic band provides a harmonic trajectory that converges to a desired target with critically damped behaviour, which is comparable to the spring used in the PMP model. The dynamic tasks use the expected principal direction of the environmental dynamics to optimise the robot posture, and the region of attraction guarantees accurate tracking advantage of non-linear FIC to compensate the perturbation. These two modules perform compensate for environmental and robot dynamics, providing an alternative to the $\lambda$-PMP model proposed in \cite{tommasino2017,tommasino2017task}. The main difference between the proposed method and the $\lambda$-PMP is that our method does not need to solve a non-linear optimisation problem in the control loop and does not require accurate knowledge of dynamics. The joint's coordination transforms this information into commands for the joints' controllers. This function is similar to what happens in the spinal cord, which is the sole reason for not implementing it directly in the region of attraction, that models the motor-cortex. Finally, the joints controller mimic the muscles behaviour using a non-linear FIC for implementing a parallel torque/position control \cite{tiseo2020bio}. 

In conclusion, the presented architecture proves that is possible to implement a hierarchical architecture of semi-autonomous controllers as theorised by \cite{ahn2012walking}, without explicitly solving an optimisation problem within the control loop. The architecture also enables the encoding of the properties of the environment dynamics, although we have not taken advantage of this property in this work. We proved that the proposed method could generate trajectories within the range of $r$-values of human movements and show behaviours consistent with the submovement theory. In future, we aim to add delays consistent with the neuronal transmission, and the controller performances during dynamic interactions with external dynamics.

\balance
\bibliography{main}
\bibliographystyle{IEEEtran}
\end{document}